\documentclass{article} % For LaTeX2e
\usepackage{iclr2021_conference,times}

\usepackage{amsmath,amsfonts,bm}

\def\eqref#1{equation~\ref{#1}}
\def\1{\bm{1}}

\DeclareMathAlphabet{\mathsfit}{\encodingdefault}{\sfdefault}{m}{sl}
\SetMathAlphabet{\mathsfit}{bold}{\encodingdefault}{\sfdefault}{bx}{n}

\usepackage{amstext,amssymb,amsfonts,latexsym}
\usepackage{algorithm,algorithmic,amsmath}
\usepackage{graphicx}

\usepackage{hyperref}
\usepackage{url}

\title{Towards Causal Federated Learning \\ For enhanced robustness and privacy}

\author
{Sreya Francis$^{1,2}$, Irene Tenison$^{1,2}$, Irina Rish$^{1,2}$\\
\normalsize{MILA$^{1}$},
\normalsize{University of Montreal$^{2}$}
}

\iclrfinalcopy % Uncomment for camera-ready version, but NOT for submission.
\begin{document}

\maketitle

\begin{abstract}
Federated Learning is an emerging privacy-preserving distributed machine learning approach to building  a shared model by performing distributed training locally on participating devices (clients) and aggregating the local models into a global one. As this approach prevents data collection and aggregation, it helps in reducing associated privacy risks to a great extent.
However, the data samples across all participating clients are
usually not independent and identically distributed (non-i.i.d.), and Out of Distribution (OOD) generalization for the learned models can be poor. Besides this challenge, federated learning also remains vulnerable to various attacks on security wherein a few malicious participating entities work towards inserting backdoors, degrading the generated aggregated model as well as inferring the data owned by participating entities. In this paper, we propose an approach for learning  invariant (causal) features common to all participating clients in a federated learning setup and analyse empirically how it enhances the Out of Distribution (OOD) accuracy as well as the privacy of the final learned model.
\end{abstract}

\section{Existing Threats}

\subsection{Domain shift issues}
While federated learning promises better privacy and efficiency, most of the existing methods ignore the fact
that the data on each client node are collected in a non-i.i.d manner, leading to data distribution shift issues between nodes \citep{Joaquin2019DomainShift}. For example, one device may take photos mostly indoors, while another mostly outdoors.
Let  $A$ and $B$ be two clients, and let $P_{A}$ and $ P_{B}$ be their associated data distributions,  respectively, in a Federated learning setup. In many real-word scenarios,    $P_{A} \neq P_{B}$.  The participating clients can have varying marginal distributions,  $\mathcal{P}_{A}(x)$, although  $\mathcal{P}(y \mid x)$ remains the same, resulting into the so-called covariate shift;  another example of shift is when marginal distributions of the class label, $\mathcal{P}_{A}(y)$, may vary across clients, even if $\mathcal{P}(x \mid y)$ is the same, and so on \citep{OpenproblemsFL2019}.

%Rewriting $P_{A}(x, y)$ as $P_{A}(y \mid x) P_{A}(x)$ and $P_{A}(x \mid y) P_{A}(y)$ allows us to characterize the differences more precisely.

\subsection{Privacy Threats}
Although the Federated Learning process makes considerable efforts to keep the user's data private,
an attacker can analyze the weights of the sent updates to make conclusions about the data of users \citep{geyer2017federated}.
Certain machine learning algorithms such as Neural Networks and Recurrent Language models are known to memorize data labelling and patterns. In such cases, a user's data may risk losing its privacy since they are represented in the model ~\citep{Mcmahan2017privatefederated}.
While this might sound unlikely if not done on purpose, there have been experiments that show it is possible to reconstruct some data points \citep{Fredrikson2015modelinversionattack}.
FL algorithms are vulnerable to some attacks, namely membership inference ~\citep{Ahmed2018}~\citep{Shokri2017inferenceattacks}, model inversion \citep{Mat2018modelinversionattack} and model extraction \citep{Florian2016stealingmodels}. 
Membership Inference typically determine whether a point is in
the training dataset or not. \citep{Shokri2017inferenceattacks} propose a shadow training technique for this attack involving training k shadow models to mimic the behavior of target model initially, then accordingly train an attack (membership inference) model. Model Inversion attacks try to use black-box access to estimate the feature values from training dataset. \citep{Mat2018modelinversionattack} explored model inversion attacks in two settings: decision trees and neural networks. 
Model Extraction attacks try to duplicate the parameters of target model. \citep{Florian2016stealingmodels} propose effective attack methods to logistic regression, neural networks and decision trees. 

\section{How can causal learning enhance federated learning}

Generalization to out-of-distribution (OOD) data in participating clients is still a challenging aspect for federated learning. This is because most statistical learning algorithms used in federated learning strongly rely on the i.i.d. assumption on client data, while in practice domain shift among participating client domains is common. As compared to associational models that are being used in federated learning, models that are learnt with respect to causal features always exhibit better generalization to non-iid data i.e. data from different distributions. 
As far as privacy is concerned, one of the main attacks posed to federated learning is membership inference attacks wherein only the model predictions can be observed by the attacker  \citep{Yeom2018generalizationvsprivacyattack},
\citep{Milad2018privacyattack}. In \citep{Milad2018privacyattack}, it has been proved that the distribution of the training data as well as the generalizability of the model significantly contribute to the membership leakage. Particularly, they show that overfitted models are more susceptible to membership inference attacks than generalized models. Hence it can be inferred that such inference attacks can be nullified to a greater extent with learning networks that exhibit better generalization. In  \citep{shruthi2020causalprivacy}, the generalization property of causal learning has been proven wherein they establish a theoretical link between causality and privacy. It is shown that models learnt using causal features generalize better to unseen data, especially on data from different distributions than the train distribution. It was also proved that causal models provide better differential privacy guarantees as compared to the current associational models that we use. 
With our approach, we explore how causal learning can enhance the out of distribution robustness as well as the impact it can have on privacy enhancement in a federated learning setup.

\section{Proposed Approach - \textbf{CausalFed}}
\label{gen_inst}
\subsection{Implementation workflow}

\begin{figure}[H]
    \centering
    \includegraphics[width=14cm, height=7cm]{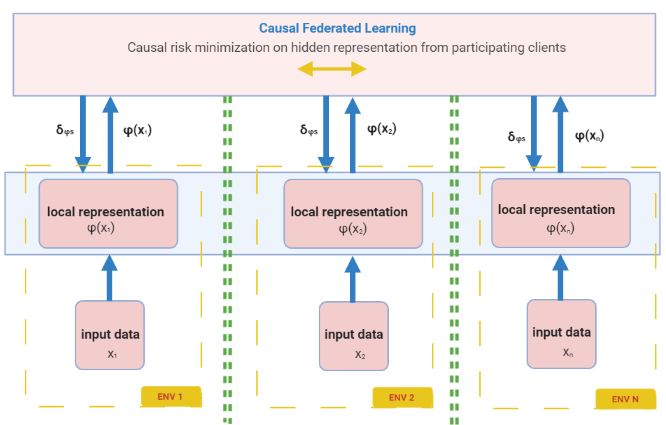}
    \caption[Causal Federated  Learning]{ Causal Federated Learning}
    \label{fig:CFed}
    \centering
\end{figure}

Keeping the data private, we propose an approach to collaboratively learn causal features common to all the participanting clients in a federated learning setup. In our federated causal learning framework, the client layer (local) is the one where in each of the participating client entities do the local training for extracting features from their respective input data and outputs the respective features in the form of numerical vectors. Consider client data $\mathcal{D}_C = {(x_i^C,y_i^C)}_{i=1}^{N_C}$
where $x_i^C$ is $i^{th}$ input and $y_i^C$ is $i^{th}$ label for client C. The hidden representation of each participating client is produced by neural network as 
$$h_i^C = \phi^C(x_i^C)$$
where $h^C \in \mathcal{R}^{N_C\times d}$
, d is the dimension of hidden representation layer.
The global server layer is for the participating clients to exchange intermediate training components and train the federated model in collaboration by minimizing the empirical
average loss as well as regularizing the model by the gradient norm of the loss for all the participating entities/environments as:

$$ \sum_{C,i}^{S, N_C}\mathcal{L}_d(w\circ h_i,y_i) + \lambda\sum_{C}^{S}{\Bigl \| \nabla_{w|w=1.0}\sum_{i}^{ N_C}\mathcal{L}_d(w\circ h_i,y_i)\Bigl \|}^2 $$

where S equals set of clients/ source domains, $N_C$ equals number of samples per client $C$, $\mathcal{L}_d$ equals classification loss, and $h$, $y$ to represent the hidden representation and its
corresponding true class label and $\lambda$ is hyperparameter.
With Invariant Risk Minimization (IRM) \citep{Martin2019irm} we attempt to learn invariant predictors in a federated learning setup that can attain an optimal empirical risk on all the participating client domains. \ref{approach2} lists an alternative approach called CausalFedGSD for the same problem.
\subsection{Algorithm}
\begin{algorithm}[H]
\caption{CausalFed}
$\textbf{ServerCausalUpdate:}
  $
  \begin{algorithmic}[H]
    
    \STATE $\text{Initialize }\mathbf{W}_{0}^{s}
    $
    
    \FOR{$\text{each server epoch, t = 1,2,..k }$}
    \STATE$\text{Select random set of S clients}$
    \STATE$\text{Share initial model with the selected clients}$
      \FOR{$\text{each client }k \in S $}
        \STATE $(\boldsymbol{\phi}(x^k_{t}), \mathbf{Y}^{k}) \leftarrow ClientRepresentation(k,\mathbf{W}_{t}^{k})$
        \STATE $ \text{Evaluate loss } \mathcal{L}_k  $
      \ENDFOR
      \STATE  $ \mathcal{L}_s = \sum_{k}^{S}\mathcal{L}_k + \lambda\sum_{k}^{S}{\Bigl \| \nabla\mathcal{L}_k\Bigl \|}^2$

      \STATE $\mathbf{W}_{t+1}^{s} \leftarrow \mathbf{W}_{t}^{s}-\eta \nabla \mathcal{L}_s$
      
    \ENDFOR
    
    \STATE $  \mathbf{W}_{t}^{k} \leftarrow  ClientUpdate(\nabla \mathcal{L}_s) $
    
  \end{algorithmic}
  
  $\textbf{ClientRepresentation($\mathbf{W}_{t}^{k}$):}
  $
  \begin{algorithmic}[H]
   \IF{\text{k is first client to start training}}
   \STATE $\mathbf{W}_{t}^{k }\leftarrow \text{initial weights from server}$
   \ELSE{}
    \STATE $\mathbf{W}_{t}^{k }\leftarrow \mathbf{W}_{t-1}^{k-1} \text{from the previous } ClientUpdate(\nabla \mathcal{L}_s)$
    
    \ENDIF

    \FOR{$\text{each local client epoch, i=1,2,..k}$}
        
        \STATE $\text{Calculate hidden representation }\boldsymbol{\phi}(x^{k}_{t}) $
    \ENDFOR
    
    \textbf{return} $ \boldsymbol{\phi}(x^{k}_{t}) \text{ and } \mathbf{Y^{k}} \text{ to server}  $
    
  \end{algorithmic}
  
  $\textbf{ClientUpdate:}$
  \begin{algorithmic}[H]
  
  \FOR{$\text{each client }k \in S $}

     \STATE $\mathbf{W}_{t+1}^{k} \leftarrow \mathbf{W}_{t}^{k}-\eta \nabla \mathcal{L}_s$
        
    \ENDFOR
    \STATE $\textbf{return }  \mathbf{W}_{t+1}^{k} \text{ to server }  $
        
  \end{algorithmic}
  
\end{algorithm}

\section{Dataset Details}
\label{headings}

\textbf{Colored MNIST}: Unlike the MNIST dataset which consists of digits 0-9 in grayscale, the colored MNIST dataset consists of input images with digits 0-4 colored red and labelled 0 while digits 5-9 are colored green with label with shape of the digit as the causal feature. In our causal federated learning setup, we split the dataset to two environments, each corresponding to a participant/client.
We sample 2000 data points per client/server domain.
Within the client environments, 80 - 90 \% of inputs have their color correlated to the digit whereas within the central server test enviroment has just 10\% color-digit correlation which helps in testing the robustness despite the spurious correlation within the inputs. 

\textbf{Rotated MNIST}: This dataset consist of original MNIST split to multiple client/participating environments by rotating each digit[0-9] with angles $0^{\circ}, 15^{\circ}, 30^{\circ}, 45^{\circ}, \text{ and } 60^{\circ}.$ We sample 1000 data points per client/server environment. The server side test domain consist of digits with angles $75^{\circ}\text{ and } 90^{\circ}$

\textbf{Rotated Fashion MNIST}: Fashion-MNIST is a dataset of Zalando's article images—consisting of a training set of 60,000 examples and a test set of 10,000 examples. Here again we split the dataset to multiple client/participating environments by rotating each fashion item with angles $0^{\circ}, 15^{\circ}, 30^{\circ}, 45^{\circ}, \text{ and } 60^{\circ}.$ We sample 10000 data points per client/server environment. The server side test domain consist of fashion items with angles $75^{\circ}\text{ and } 90^{\circ}$

\section{Results}
\label{others}

In our experiments, we compare the performance of federated averaging (Fed-Avg) with the following approaches:

\textbf{Fed-ERM}
Within the CausalFed setup, this approach minimizes the empirical average of loss over training data points and treats the data from different domains as i.i.d.
ERM loss is given by:
$$ \sum_{C,i}^{S, N_C}\mathcal{L}_s(w\circ h_i,y_i)$$
where S equals set of clients/ source domains, $N_C$ equals number of samples per client $C$, $\mathcal{L}_s$ equals classification loss. \\
\textbf{CausalFed-RM}
In this approach, we minimize the random match(RMatch) causal loss \citep{Divyat2020} within the CausalFed setup. RMatch loss is given by:
$$ \sum_{C,i}^{S, N_C}\mathcal{L}_s(w\circ h_i,y_i)+\lambda *\sum_{\Omega(j, k)=1 \mid j \sim N_{C}, k \sim N_{C^{\prime}}} \operatorname{Dist}(h_{j}, h_{k})$$ where $\Omega$ represents the match function used to randomly pair the data points across the different client domains.\\
\textbf{CausalFed-IRM}
In this approach, we minimize the IRM loss \citep{Martin2019irm} within the CausalFed setup.

\begin{table}[H]
\caption{Train Results}
\label{table4}
\begin{center}
\begin{tabular}{llllll}
\hline 
\multicolumn{1}{c}{\bf Dataset}  &\multicolumn{1}{c}{\bf Arch} &\multicolumn{1}{c}{\bf Fed-Avg }&\multicolumn{1}{c}{\bf Fed-ERM } &\multicolumn{1}{c}{\bf CausalFed-RM } &\multicolumn{1}{c}{\bf CausalFed-IRM }
\\ \hline 
Colored MNIST         &ResNet18 &80.3\% &\textbf{82.97} \%   &60.42 \%     &59.33 \%  \\ \hline
Rotated MNIST         &ResNet18 &85.2\%  &\textbf{86.5} \%     &79.8 \%    &80.2 \%   \\ \hline 
Rotated FMNIST         &LeNet &81.4\%   &\textbf{82.3} \%     &72.1 \%       &71.5 \%  \\ \hline 

\end{tabular}
\end{center}
\end{table}

\begin{table}[H]
\caption{Test Results}
\label{table5}
\begin{center}
\begin{tabular}{llllll}
\hline 
\multicolumn{1}{c}{\bf Dataset}  &\multicolumn{1}{c}{\bf Arch} &\multicolumn{1}{c}{\bf Fed-Avg }&\multicolumn{1}{c}{\bf Fed-ERM } &\multicolumn{1}{c}{\bf CausalFed-RM } &\multicolumn{1}{c}{\bf CausalFed-IRM }
\\ \hline 
Colored MNIST         &ResNet18 &11\% &10.2 \%   &\textbf{65.62} \%     &60.3 \%  \\ \hline
Rotated MNIST         &ResNet18 &82.7\%  &82.9 \%     &\textbf{90.2} \%    &89.1 \%   \\ \hline 
Rotated FMNIST         &LeNet   &72\% &71.6 \%     &\textbf{74.6} \%     &73.9 \%  \\ \hline 

\end{tabular}
\end{center}
\end{table}
We observed that when clients have out of distribution data in a federated setup, FedAvg as well as FedERM does not fare well in the server side test data set though they give highly accurate results on train data(iid) whereas CausalFed-RM and CausalFed-IRM perfoms much better on test data(non iid).

\textbf{Privacy Leakage} In our experiments, within the CausalFed setup, we analyse the privacy leakage on 3 common attacks namely, Membership inference attack, Property inference attack and Backdoor attack. The privacy leakage on each of the attacks is measured by testing the accuracy of attack model. Details on each of the attacks are added in \ref{attack} \ref{backdoor}.
\begin{table}[H]
\caption{Leakage on inference attack}
\label{table6}
\begin{center}
\begin{tabular}{lllll}
\hline 
\multicolumn{1}{c}{\bf Dataset}   &\multicolumn{1}{c}{\bf Fed-Avg } &\multicolumn{1}{c}{\bf Fed-ERM } &\multicolumn{1}{c}{\bf CausalFed-RM } &\multicolumn{1}{c}{\bf CausalFed-IRM }
\\ \hline 
Colored MNIST     &79.21 \%    &79.45 \%   &58.57 \%     &56.9 \%  \\ \hline
Rotated MNIST     &84.4 \%     &85.24 \%     &68.3 \%    &64.4 \%   \\ \hline 
Rotated FMNIST    &76.61 \%     &78.23 \%     &57.55 \%       &55.7 \%  \\ \hline 

\end{tabular}
\end{center}
\end{table}

We observe that in our setup with an out of distribution(OOD) test set, the membership inference attack accuracy of a federated causal client adversary model is much lesser as compared to a federated setup with associational client models. It was also observed that federated causal models provide better pivacy guarantees against property inference attacks which could be owed to the fact that inversion based on learning correlations between attributes and final prediction, e.g., using color to predict the digit, can be eliminated by causal models, since a non-causal feature will not be included in our final causal federated model.

\section{Conclusion}
In this work, we show that CausalFed is more accurate than non-privacy-preserving federated learning approaches as well as  superior to non-federated associational learning approaches in comparison to existing privacy enhancing approaches in federated setup which suffer from pretty high accuracy loss. We were able to experiment and confirm that causal feature learning can enhance out of distribution robustness in federated learning. Moving forward, we need to analyse the performance of this approach in real world datsets as well as compare various other causal learning approaches which can further enhance the out of distribution robustness and improve leakage protection in our current setup. We believe that CausalFed and CausalFedGSD serve as an initial approach to perform causal learning in a federated setting that offers several extensions for future work.

\bibliography{iclr2021_conference}

\begin{thebibliography}{17}
\providecommand{\natexlab}[1]{#1}
\providecommand{\url}[1]{\texttt{#1}}
\expandafter\ifx\csname urlstyle\endcsname\relax
  \providecommand{\doi}[1]{doi: #1}\else
  \providecommand{\doi}{doi: \begingroup \urlstyle{rm}\Url}\fi

\bibitem[Arjovsky et~al.(2019)Arjovsky, Bottou, Gulrajani, and
  Paz]{Martin2019irm}
Martin Arjovsky, Léon Bottou, Ishaan Gulrajani, and David~Lopez Paz.
\newblock Invariant risk minimization.
\newblock \emph{arXiv:1907.02893}, 2019.

\bibitem[C et~al.(2019)C, E, Cristofaro, and Shmatikov]{Melis2018}
Melis C, Song E, De~Cristofaro, and V.~Shmatikov.
\newblock Exploiting unintended feature leakage in collaborative learning.
\newblock \emph{IEEE Symposium on Security and Privacy}, 2019.

\bibitem[Fredrikson et~al.(2015)Fredrikson, S, and
  T]{Fredrikson2015modelinversionattack}
Matt Fredrikson, Jha S, and Ristenpart T.
\newblock Model inversion attacks that exploit confidence information and basic
  countermeasures.
\newblock \emph{ACM}, pp.\  pp. 1322--1333, 2015.

\bibitem[Fredrikson et~al.(2018)Fredrikson, Jha, and
  Ristenpart]{Mat2018modelinversionattack}
Matt Fredrikson, Somesh Jha, and Thomas Ristenpart.
\newblock Model inversion attacks that exploit confidence information and basic
  countermeasures.
\newblock \emph{arXiv:1806.01246}, 2018.

\bibitem[Geyer et~al.(2017)Geyer, Klein, and Nabi]{geyer2017federated}
Robin~C. Geyer, Tassilo Klein, and Moin Nabi.
\newblock Differentially private federated learning: A client level
  perspective.
\newblock \emph{arXiv:1712.07557}, 2017.

\bibitem[Kairouz et~al.(2019)Kairouz, Mcmahan, Avent, Bellet, Bennis, Bhagoji,
  Bonawitz, Charles, Cormode, and Cummings]{OpenproblemsFL2019}
Peter Kairouz, Brendan Mcmahan, Brendan Avent, Aurélien Bellet, Mehdi Bennis,
  Arjun~Nitin Bhagoji, Keith Bonawitz, Zachary Charles, Graham Cormode, and
  Rachel Cummings.
\newblock Advances and open problems in federated learning.
\newblock \emph{arxiv}, 2019.

\bibitem[Mahajan et~al.(2020)Mahajan, Tople, and Sharma]{Divyat2020}
D~Mahajan, S~Tople, and A~Sharma.
\newblock Domain generalization using causal matching.
\newblock \emph{arXiv:2006.07500}, 2020.

\bibitem[McMahan et~al.(2017)McMahan, D.Talwar, and
  K.Zhang]{Mcmahan2017privatefederated}
H~Brendan McMahan, Ramage D.Talwar, and K.Zhang.
\newblock Learning differentially private language models without losing
  accuracy.
\newblock \emph{arXiv:1710.06963}, 2017.

\bibitem[Nasr et~al.(2018{\natexlab{a}})Nasr, Shokri, and Houmansadr]{Nasr2018}
Nasr, Reza Shokri, and Amir Houmansadr.
\newblock Comprehensive privacy analysis of deep learning: Passive and active
  white-box inference attacks against centralized and federated learning.
\newblock \emph{arXiv:1812.00910}, 2018{\natexlab{a}}.

\bibitem[Nasr et~al.(2018{\natexlab{b}})Nasr, Shokri, and
  Houmansadr]{Milad2018privacyattack}
Milad Nasr, Reza Shokri, and Amir Houmansadr.
\newblock Comprehensive privacy analysis of deep learning.
\newblock \emph{arXiv:1812.00910}, 2018{\natexlab{b}}.

\bibitem[Quionero-Candela et~al.(2019)Quionero-Candela, Sugiyama, Schwaighofer,
  , and Lawrence]{Joaquin2019DomainShift}
Joaquin Quionero-Candela, Masashi Sugiyama, Anton Schwaighofer, , and Neil~D
  Lawrence.
\newblock Dataset shift in machine learning.
\newblock \emph{ISBN 0262170051}, 2019.

\bibitem[Salem et~al.(2018)Salem, Zhang, Humbert, Berrang, Fritz, and
  Backes]{Ahmed2018}
Ahmed Salem, Yang Zhang, Mathias Humbert, Pascal Berrang, Mario Fritz, and
  Michael Backes.
\newblock Mlleaks model and data independent membership inference attacks and
  defenses.
\newblock \emph{arXiv:1806.01246}, 2018.

\bibitem[Shokri et~al.(2017)Shokri, Stronati, Song, and
  Shmatikov]{Shokri2017inferenceattacks}
Reza Shokri, Marco Stronati, Congzheng Song, and Vitaly Shmatikov.
\newblock Membership inference attacks against machine learning models.
\newblock \emph{arXiv:1806.01245}, 2017.

\bibitem[Tople et~al.(2020)Tople, Sharma, and Noris]{shruthi2020causalprivacy}
Shruti Tople, Amit Sharma, and Aditya~V. Noris.
\newblock Alleviating privacy attacks via causal learning.
\newblock \emph{arXiv:1909.12732}, 2020.

\bibitem[Tramèr et~al.(2016)Tramèr, Zhang, Juels, Reiter, , and
  Ristenpart]{Florian2016stealingmodels}
Florian Tramèr, Fan Zhang, Ari Juels, Michael~K Reiter, , and Thomas
  Ristenpart.
\newblock Stealing machine learning models via prediction apis.
\newblock \emph{arXiv:1806.01246}, 2016.

\bibitem[Yeom et~al.(2018)Yeom, Giacomelli, Fredrikson, and
  Jh]{Yeom2018generalizationvsprivacyattack}
S.~Yeom, I.~Giacomelli, M.~Fredrikson, and S.~Jh.
\newblock Privacy risk in machine learning: Analyzing the connection to
  overfitting.
\newblock \emph{arXiv:1709.01604}, 2018.

\bibitem[Zhao et~al.(2018)Zhao, Li, Lai, Suda, Civin, and
  Chandra]{Yue2018fednoniid}
Yue Zhao, Meng Li, Liangzhen Lai, Naveen Suda, Damon Civin, and Vikas Chandra.
\newblock Federated learning with non-iid data.
\newblock \emph{arXiv:1806.00582}, 2018.

\end{thebibliography}
\bibliographystyle{iclr2021_conference}

\appendix
\section{Appendix}
\subsection{Objective Functions}\label{obj}
\textbf{ERM}
This approach minimizes the empirical average of loss over training data points and treats the data from different domains as i.i.d.
ERM loss is given by:
$$ \sum_{C,i}^{S, N_C}\mathcal{L}_s(w\circ h_i,y_i)$$
where S equals set of clients/ source domains, $N_C$ equals number of samples per client $C$, $\mathcal{L}_s$ equals classification loss. \\
\textbf{RMatch}
RMatch loss is given by:
$$ \sum_{C,i}^{S, N_C}\mathcal{L}_s(w\circ h_i,y_i)+\lambda *\sum_{\Omega(j, k)=1 \mid j \sim N_{C}, k \sim N_{C^{\prime}}} \operatorname{Dist}(h_{j}, h_{k})$$ where $\Omega$ represents the match function used to randomly pair the data points across the different client domains \citep{Divyat2020}.

\subsection{Inference Attack}\label{attack}

\subsubsection{Membership Inference}

The main idea is that each training data point affects the gradients of the loss function such that the adversary can use Stochastic
Gradient Descent algorithm (SGD) to extract information from
other clients’ data \citep{Nasr2018}. The adversary can perform
gradient ascent on a target data point before local parameter update. SGD reduces the gradient,in case the considered data point is part of a client’s set resulting in a succesful membership inference. Attack can come from both the client side and the server side. An adversarial client can observe the aggregated model updates and extract information about the union of the training dataset of all other participants by injecting adversarial model updates.
For a server side attack, it can control the view of each target participant on the aggregated model updates and extract information
from its dataset.

In our implementation, we sample 2,000 datapoints
for Rotated-MNIST and 10,000 datapoints for Fashion-MNIST from the original train and test dataset to create the attack-train and attack-test dataset. We use pytorch code provided by \citep{Shokri2017inferenceattacks}\citep{Nasr2018}

\subsubsection{Property Inference}

The main idea behind this attack is that, at each round,
each client’s contribution is based on a batch of their local
training data, so the attacker can infer properties that characterize the target dataset for which the adversary needs sample train data, which is labeled with the attribute to be infered.\citep{Melis2018} It is aimed at 
infering properties of client data that are uncorrelated with the
features that characterize the classes of the model. In our experiments we decided on client domain as the attribute which is to be inferred by the adversary. Another such attribute that is uncorrelated with the final prediction is the color of the input.

We observe that federated causal models provide better pivacy guarantees against this attack which could be owed to the fact that inversion based on learning correlations between
attributes and final prediction, e.g., using color to predict the digit, can be eliminated by causal models, since a
non-causal feature will not be included in the our final causal federated model.

\subsection{Backdoor Attack}\label{backdoor}
For an initial analysis, we experimented with two backdoor attacks:
\begin{itemize}
    
    \item A single-pixel attack, where in the attacker changes the top-left pixel color of all the inputs, and mislabels them.
    \item A semantic backdoor where in the attacker selects certain
    features as the backdoors and misclassifies them. For example, the attacker classifies digits rotated $15^{\circ}$ with label 7 as 0
\end{itemize}
In both the cases, CausalFed exhibited better resilience as compared to FedAvg.
\subsection{Network Architecture}

\begin{table}[H]
\caption{Network Architecture}
\label{table1}
\begin{center}
\begin{tabular}{lllll}
\hline 
\multicolumn{1}{c}{\bf Architecture}  &\multicolumn{1}{c}{\bf No of Layers} &\multicolumn{1}{c}{\bf Kernel spec} 
\\ \hline 
LeNet         &5  &(5x5), (2x2)   
\\ \hline
AlexNet         &8   &(11x11), (5x5), (3x3)       
\\ \hline 
ResNet18         &18    &(7x7), (3x3)      
\\ \hline 

\end{tabular}
\end{center}
\end{table}

\subsection{CasualFedGSD - Alternative approach to CausalFed} \label{approach2}
\begin{figure}[H]
    \centering
    \includegraphics[width=8cm, height=5cm]{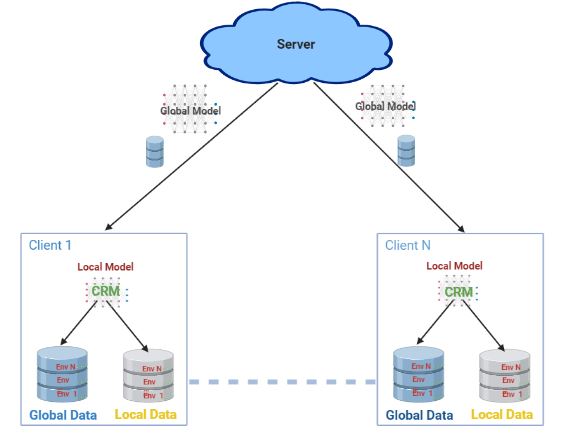}
    \caption[Causal Federated  Learning]{ Causal Federated Learning}
    \label{fig:GDS}
    \centering
\end{figure}

In ~\citep{Yue2018fednoniid}, it has been shown that globally shared data can reduce EMD(earth mover's distance)  between the data distribution on clients and the population distribution which can help in improved test accuracy.As this globally shared data is a separate dataset from that of the client, this approach is not privacy sensitive.

With the CausalFed approach, there can be privacy concerns regarding sharing the client data representation to the global server due to which depending on a global data set(with different enviroments) to enhance causal feature learning within a federated learning setup seems plausible.
 As we have no control on the clients’ data, we can distribute a small subset of global data containing a distribution over all the classes/enviroments from the server side to the clients during the initialization stage of federated learning. 
 
 The local model of each client is learned by minimizing the empirical average loss as well as regularizing the model by the gradient norm of the loss for both the shared data from server(Global Environment) and private data from each client(Local Environment). This enhances the learning of causal/invariant features common to both the client and global data environments without losing the privacy of client side data.
 
 \begin{algorithm}[H]
\caption{CausalFedGSD}
$\textbf{ServerUpdate:}
  $
  \begin{algorithmic}[H]
    \STATE $ \text{G } \leftarrow \text{distribution over all environments present in server}$
    \STATE $\text{Initialize }w_0
    $
    \STATE $\text{Initialize random portion of G as } G_0$
    \FOR{$\text{each server epoch, t = 1,2,..k }$}
    \STATE$\text{Select random set of S clients}$
    \STATE$\text{Share $G_0$ and initial model with the selected clients}$
      \FOR{$\text{each client }k \in S $}
        \STATE $w^k_{t+1} = ClientUpdate(k,w_{t})
        $
      \ENDFOR
      \STATE $w_{t+1} = \sum_{k=1}^{K}\frac{n_k}{n}w_{t+1}^k
      $
    \ENDFOR
    
  \end{algorithmic}
  $\textbf{ClientUpdate(w):}
  $
  \begin{algorithmic}[H]
    \STATE $\mathcal{E}_{\mathrm{tr}} \in [\text{Client Env}] \bigcup [\text{Global Env}] 
    $
    \FOR{$\text{each local client epoch, t=1,2,..k}$}
      \STATE $
L_{\mathrm{IRM}}(\Phi, w^k_{t})=\sum_{e \in \mathcal{E}_{\mathrm{tr}}} R^{e}(w \circ \Phi)+\lambda \cdot \mathbb{D}(w, \Phi, e)
 $
      \STATE $w^k_{t} = w^k_{t} - \eta\nabla L_{\mathrm{IRM}}(w^k_{t})
      $
    \ENDFOR
    
    $\text{return w to server}$
  \end{algorithmic}
\end{algorithm}

 \begin{table}[H]
\caption{Train Results}
\label{table2gsd1}
\begin{center}
\begin{tabular}{llllll}
\hline 
\multicolumn{1}{c}{\bf Dataset}  &\multicolumn{1}{c}{\bf Arch} &\multicolumn{1}{c}{\bf Fed-Avg }&\multicolumn{1}{c}{\bf Fed-ERM } &\multicolumn{1}{c}{\bf CausalFedGSD-RM } &\multicolumn{1}{c}{\bf CausalFedGSD-IRM }
\\ \hline 
Colored MNIST         &ResNet18 &80.3\% &\textbf{82.97} \%   &57.42 \%     &55.32 \%  \\ \hline
Rotated MNIST         &ResNet18 &85.2\%  &\textbf{86.5} \%     &73.7 \%    &77.2 \%   \\ \hline 
Rotated FMNIST         &LeNet &81.4\%   &\textbf{82.3} \%     &69.2 \%       &68.6 \%  \\ \hline 

\end{tabular}
\end{center}
\end{table}

\begin{table}[H]
\caption{Test Results}
\label{table3gsd2}
\begin{center}
\begin{tabular}{llllll}
\hline 
\multicolumn{1}{c}{\bf Dataset}  &\multicolumn{1}{c}{\bf Arch} &\multicolumn{1}{c}{\bf Fed-Avg }&\multicolumn{1}{c}{\bf Fed-ERM } &\multicolumn{1}{c}{\bf CausalFedGSD-RM } &\multicolumn{1}{c}{\bf CausalFedGSD-IRM }
\\ \hline 
Colored MNIST         &ResNet18 &11\% &10.2 \%   &\textbf{55.62} \%     &52.3 \%  \\ \hline
Rotated MNIST         &ResNet18 &82.7\%  &82.9 \%     &\textbf{85.2} \%    &83.1 \%   \\ \hline 
Rotated FMNIST         &LeNet   &72\% &71.6 \%     &\textbf{71.9} \%     &70.2 \%  \\ \hline 

\end{tabular}
\end{center}
\end{table}

\end{document}